%% file: root.tex
\documentclass[letterpaper, 10 pt, conference]{ieeeconf}
\IEEEoverridecommandlockouts
\title{\LARGE \bf
{GSplatLoc: Grounding Keypoint Descriptors into 3D Gaussian Splatting for Improved Visual Localization\textsuperscript{†}\thanks{\textsuperscript{†}This work has been submitted to the IEEE for possible publication. Copyright may be transferred without notice, after which this version may no longer be accessible.}} 
}

\author{Gennady Sidorov $^{1, 2}$, Malik Mohrat $^{1, 2}$, Denis Gridusov$^{1}$, Ruslan Rakhimov $^{3}$, and Sergey Kolyubin$^{1}$
\thanks{$^{1}$ BE2R Lab, ITMO University, St. Petersburg, Russia. {\tt\small \{gksidorov, mmohrat, ddgridusov, s.kolyubin\}@itmo.ru}}%
\thanks{$^{2}$ Robotics Center, Moscow, Russia.}
\thanks{$^{3}$ T-Tech, Moscow, Russia.}
}

\input{preamble}

\begin{document}




\maketitle


\input{src/00-abstract}
\input{src/01-introduction}
\input{src/02-related-work}
\input{src/03_1-Preliminaries}
\input{src/03-method}

\input{src/04-experiments}
\input{src/05-conclusion}


\addtolength{\textheight}{-5.2cm}   








                                  
\bibliographystyle{IEEEtran}
\bibliography{IEEEabrv,main}

\end{document}

%% file: preamble.tex
\usepackage{caption}
\usepackage{amsmath}

\usepackage[T1]{fontenc}
\usepackage[accsupp]{axessibility}  

\usepackage{hyperref}
\hypersetup{
    colorlinks=true,
    linkcolor=blue,
    filecolor=magenta,      
    urlcolor=cyan,
    pdftitle={GSplatLoc},
    pdfpagemode=FullScreen,
    }

\usepackage{lipsum}

\usepackage{booktabs}

\usepackage{graphicx} 
\usepackage{subcaption}
\usepackage{tabularx}
\usepackage{multirow}
\usepackage{url}
\usepackage{threeparttable}
\usepackage[table,xcdraw]{xcolor}

\newcommand{\mycomment}[1]{}

\usepackage{float} 
\definecolor{myGold}{rgb}{1.0, .875, .0}
\definecolor{mySilver}{rgb}{0.647 .663, .706}
\definecolor{myPurple}{rgb}{0.4, .0, .8}
\definecolor{myGreen}{rgb}{0, .8, .3}
\definecolor{myRed}{rgb}{0.8, .2, .2}
\definecolor{myBlue}{rgb}{0.0, .0, .8}
\definecolor{springgreen}{RGB}{204, 255, 51}
\definecolor{myWhite}{RGB}{255, 255, 255}
\colorlet{colorFst}{myGold!45}       
\colorlet{colorSnd}{mySilver!45} 
\colorlet{colorblank}{myWhite}

\newcommand{\fs}[1]{\colorbox{colorFst}{\textbf{#1}}}
\newcommand{\nd}[1]{\colorbox{colorSnd}{\textbf{#1}}}     

%% file: src/00-abstract.tex
\begin{abstract}

Although various visual localization approaches exist, such as scene coordinate regression and camera pose regression, these methods often struggle with optimization complexity or limited accuracy. To address these challenges, we explore the use of novel view synthesis techniques, particularly 3D Gaussian Splatting (3DGS), which enables the compact encoding of both 3D geometry and scene appearance. We propose a two-stage procedure that integrates dense and robust keypoint descriptors from the lightweight XFeat feature extractor into 3DGS, enhancing performance in both indoor and outdoor environments. The coarse pose estimates are directly obtained via 2D-3D correspondences between the 3DGS representation and query image descriptors. In the second stage, the initial pose estimate is refined by minimizing the rendering-based photometric warp loss. Benchmarking on widely used indoor and outdoor datasets demonstrates improvements over recent neural rendering-based localization methods, such as NeRFMatch and PNeRFLoc. Project page: \href{https://gsplatloc.github.io}{https://gsplatloc.github.io}

\end{abstract}

%% file: src/01-introduction.tex
\section{Introduction}

Visual localization is a key task in computer vision that estimates the pose of a moving camera relative to a pre-built environment map. It is a crucial function for mobile and humanoid robots navigating indoor and outdoor environments, enabling autonomous navigation and interaction within the environment. Moreover, this capability is required for robots to perceive their position in the 3D environment, which constitutes one of the core components of SLAM systems. Among various localization approaches, vision-based methods have received significant attention due to the widespread availability and low cost of cameras. Equipped with vision sensors, these methods can be implemented with minimal modification, making them highly adaptable for real-world robotic applications ~\cite{5459273, heng2019projectautovisionlocalization3d}.

Early methods for visual re-localization primarily relied on image retrieval techniques, where a query image was compared to a database of images with known poses to infer an approximate location. Although computationally efficient, these methods face scalability challenges and exhibit reduced accuracy in dynamic environments.

Structured feature-matching approaches enhance localization by leveraging 3D point clouds generated through Structure-from-Motion (SfM). These methods establish explicit 2D-3D correspondences between features extracted from query images and the 3D map, typically using Perspective-n-Point (PnP) solvers with RANSAC. Although these approaches offer high localization accuracy, they impose significant memory and computational requirements when dealing with large-scale environments~\cite{6247782}.

\begin{figure}[t]%
\includegraphics[width=0.5\textwidth]{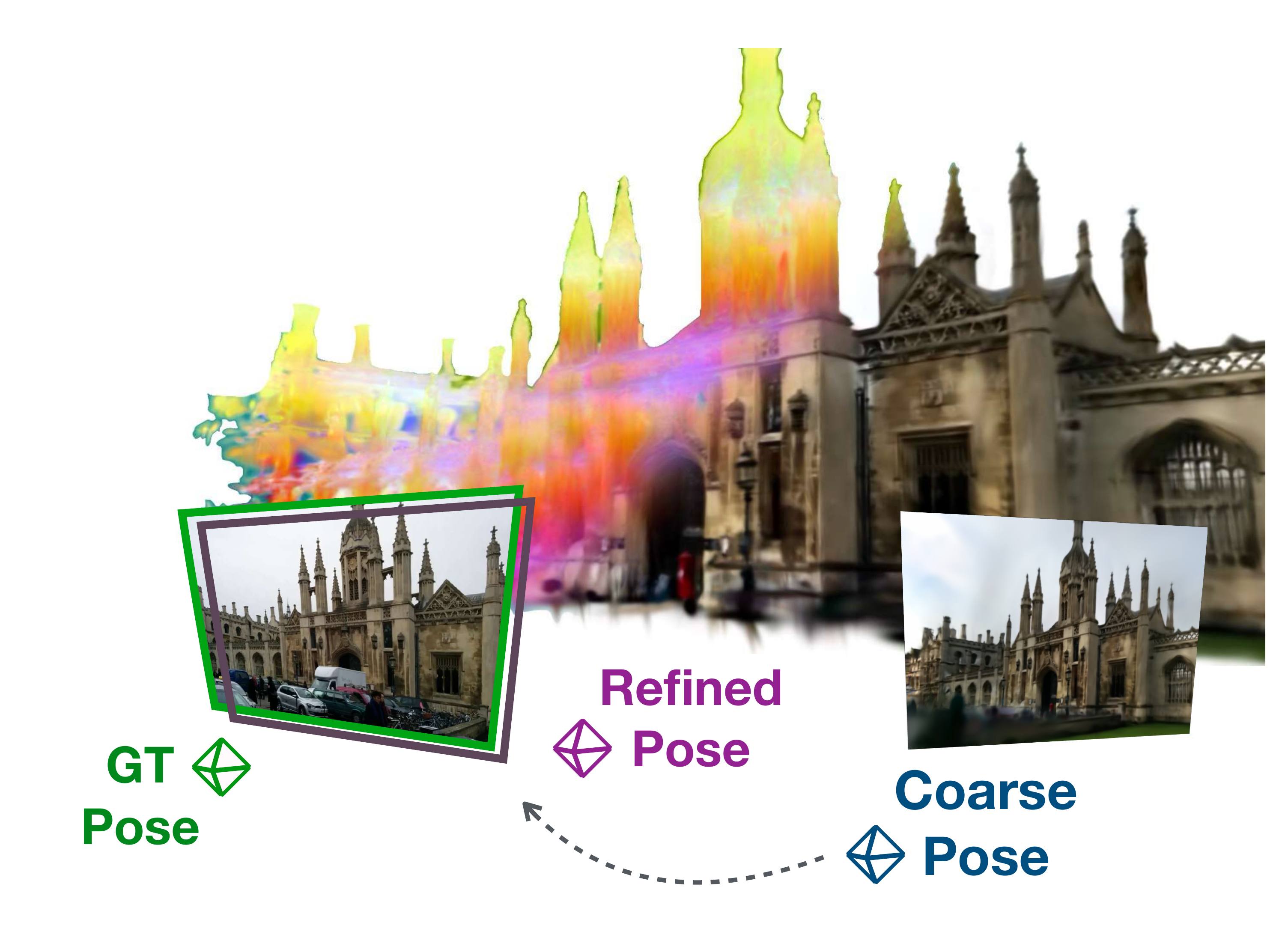}\caption{\emph{GSplatLoc} constructs a 3D Gaussian Splatting (3DGS) model with distilled descriptor features. For localization, the initial coarse pose is estimated through structural matching with these features and refined during test-time optimization using rendering-based photometric warp loss to enhance accuracy.}
\label{fig:teaser}
\end{figure}

Pose regression methods employ deep neural networks to estimate camera poses directly from input images, reducing dependence on large-scale 3D maps. These methods integrate the environmental structure into the network’s architecture, enabling end-to-end training. However, they often underperform compared to structure-based approaches due to limited generalization capabilities~\cite{sarlin21pixloc}. Absolute Pose Regression (APR) techniques provide a trade-off between computational efficiency and accuracy, while Scene Coordinate Regression (SCR) further improves performance by learning compact scene representations. However, SCR methods demand extensive optimization to achieve state-of-the-art accuracy, which constrains their applicability in real-time scenarios.

Recent advancements in neural 3D scene representations, such as Neural Radiance Fields (NeRF)~\cite{mildenhall2020nerf} and 3D Gaussian Splatting (3DGS)~\cite{kerbl20233d}, have enabled high-fidelity view synthesis. These methods enhance visual localization and contribute to Neural Render Pose (NRP) estimation by generating novel perspectives, enabling data augmentation during training and improving robust feature matching. NeRF-based methods, while effective, often suffer from slow inference speeds, long training time, and artifacts in rendered images. Enhancements such as incorporating feature fields into NeRF representations have improved robustness against such artifacts and enhanced pose estimation accuracy~\cite{tschernezki2022neural, kobayashi2022decomposing, zhou2023feature, chen2024neural}. On the other hand, 3DGS-based pipelines, such as GS-CPR~\cite{liu2025gscpr} and HGSLoc~\cite{niu2024hgsloc} leverage this 3D representation as a test-time pose refinement framework, yet they depend on external pose estimators for initialization.

To address these limitations, we propose \textbf{GSplatLoc}, a novel visual localization framework that integrates structure-based coarse pose estimation with photometric rendering-based optimization in a unified, end-to-end pipeline. Our approach leverages the point-based physically-consistent 3D Gaussian representation to distill scene-agnostic feature descriptors, enabling efficient initial pose estimation and subsequent refinement. (see Figure \ref{fig:teaser}).

Our main contributions are:
\begin{itemize}
\item A novel visual localization framework that combines structure-based keypoint matching with rendering-based pose refinement in a \emph{unified} 3D Gaussian Splatting (3DGS) pipeline.
\item An efficient test-time pose refinement strategy that leverages fast differentiable rendering and photometric warping loss minimization to improve localization accuracy, particularly in outdoor and dynamic scenes.
\item A demonstration of state-of-the-art performance on indoor and outdoor benchmarks, outperforming other NRP approaches based on NeRF, while maintaining lower computational overhead and requiring a single RGB modality.
\end{itemize}

%% file: src/02-related-work.tex
\section{Related Work}

\textbf{Absolute Pose Regression (APR)} methods directly regress camera poses from query images using deep neural networks~\cite{kendall2015posenet, brachmann2016uncertainty, brahmbhatt2017mapnet, wang2020atloc, hu2020dasgil, arnold2022map, chen2022dfnet, shavit2022camera}. PoseNet~\cite{kendall2015posenet} introduced this approach by employing a pre-trained GoogLeNet for feature extraction. Subsequent studies enhanced APR by incorporating additional modules: MapNet~\cite{brahmbhatt2017mapnet} jointly estimates both absolute and relative poses, while AtLoc~\cite{wang2020atloc} leverages self-attention to extract salient features. Recently, MaRePo~\cite{chen2024maprelativeposeregressionvisual} introduced a two-stage approach, first regressing scene-specific geometry and then refining the pose with a transformer. Although APR methods are computationally efficient, their accuracy remains inferior to that of structure-based approaches.

\textbf{Scene Coordinate Regression (SCR)} methods predict dense 2D-3D correspondences by regressing the 3D scene coordinates directly from images~\cite{shotton2013scene, brachmann2018dsac, brachmann2018learning, li2018scene, brachmann2021visual}. Early SCR approaches relied on random forests~\cite{shotton2013scene}, whereas more recent methods leverage deep learning. DSAC~\cite{brachmann2018dsac} introduced differentiable RANSAC for end-to-end training, later extended by DSAC++~\cite{brachmann2018learning} and ESAC~\cite{brachmann2019expert}, which utilize a gating network to divide the problem into simpler sub-tasks. ACE~\cite{brachmann2023ace} and GLACE~\cite{wang2024glacegloballocalaccelerated} improve efficiency by avoiding end-to-end supervision and leveraging shuffled pixel-based training. Although SCR methods achieve higher accuracy, they necessitate high-quality 3D models and extensive training data.

\textbf{Neural Render Pose (NRP)} estimation utilizes neural rendering techniques such as NeRF~\cite{mildenhall2020nerf} and 3D Gaussian Splatting (3DGS)~\cite{kerbl20233d} to improve localization by synthesizing novel views and refining camera poses. These methods refine pose estimates by minimizing the photometric or feature-based discrepancies between observed and rendered images. iNeRF~\cite{yenchen2021inerf} employs inverse rendering to iteratively refine poses, while LENS~\cite{moreau2021lens} generates synthetic training data using NeRF-W~\cite{nerfw}. DFNet~\cite{chen2022dfnet} enhances pose estimation through direct feature matching between query and rendered images, and PNeRFLoc~\cite{zhao2023pnerfloc} aligns 2D-3D correspondences via NeRF-based feature warping. Recent approaches~\cite{kobayashi2022decomposing, tschernezki2022neural, zhou2024nerfectmatchexploringnerf} integrate learned feature fields to enhance localization; however, they incur high computational costs due to extended training and rendering times. In contrast, our method exploits the efficiency of 3DGS for rapid rendering and distills feature fields to refine camera pose estimates using only RGB data.

\textbf{Keypoint detection and descriptor learning} have evolved through deep learning, enhancing feature matching for localization. CNN-based methods, including SuperPoint~\cite{detone2018superpoint}, D2-Net~\cite{dusmanu2019d2net}, and R2D2~\cite{revaud2019r2d2}, improve keypoint extraction and description. Transformer-based models like LoFTR~\cite{sun2021loftr} and LightGlue~\cite{lindenberger2023lightglue} further refine feature matching by capturing long-range dependencies. Recent methods such as XFeat~\cite{potje2024cvpr} offer lightweight feature extraction for improved efficiency. Feature matching-based localization methods ~\cite{lindenberger2021pixelperfect, chen2024neural, chen2022dfnet} employ deep descriptors to enhance pose refinement robustness. Our method integrates keypoint-based descriptors with 3DGS feature fields to improve pose estimation accuracy.

\begin{figure*}[t]
\centering
\includegraphics[width=0.99\textwidth]{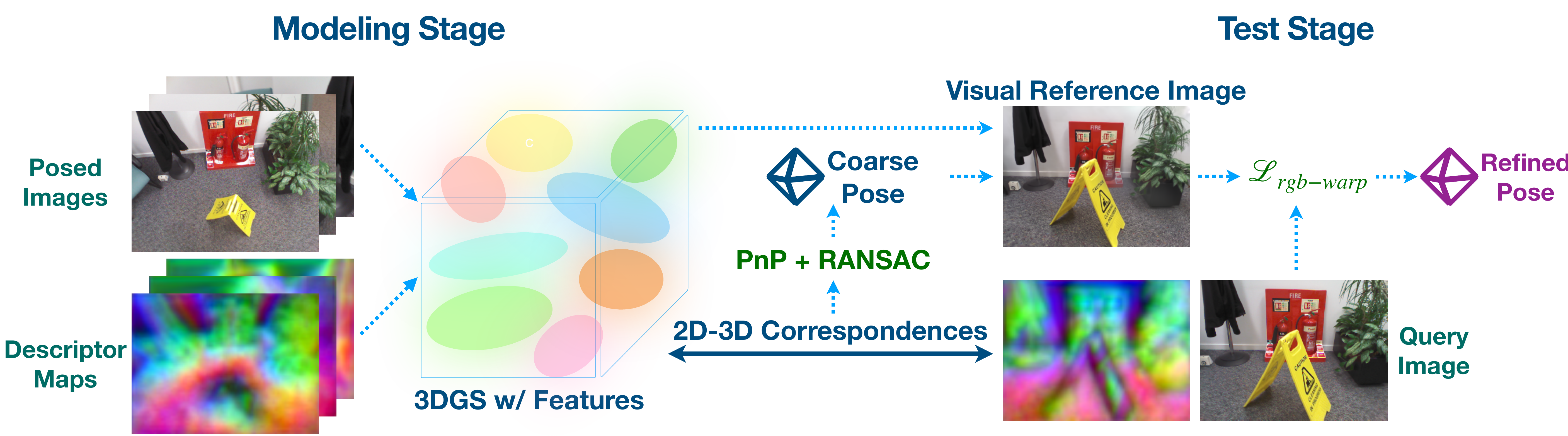}
\caption{\emph{Overview of the GSplatLoc Base pipeline.} First, we model the scene using a feature-based 3D Gaussian Splatting (3DGS) approach, leveraging the XFeat~\cite{potje2024cvpr} network for feature extraction and distillation. In the test stage, the initial coarse pose is estimated by matching 2D keypoints from the query image to 3D features in the 3DGS model, which is then refined using a Perspective-n-Point (PnP) solver within a RANSAC loop. We then refine the coarse pose by aligning the image rendered from 3DGS with the input query image using an RGB warping loss. This process enhances pose accuracy via test-time optimization.}
\label{fig:pipeline}
\end{figure*}

%% file: src/03_1-Preliminaries.tex
\section{Preliminaries}

Our method integrates keypoint descriptor models with 3D Gaussian Splatting, embedding keypoint features into the 3D representation to enhance re-localization accuracy. We provide a concise overview of this process.

Building on the Feature-3DGS framework~\cite{zhou2023feature}, we employ a modified 3DGS to distill high-dimensional features into a feature field while simultaneously constructing a radiance field. This approach utilizes N-dimensional parallel Gaussian rasterization to accelerate computation, facilitating seamless integration with various 2D foundation models.

We initialize the 3D Gaussians using a point cloud obtained through Structure-from-Motion (SfM) reconstruction. Their projection into 2D space involves transforming covariance matrices and incorporating rotation, scaling, opacity, spherical harmonics, color, and other visual features.

Pixel colors and feature values are computed via $\alpha$-blending under the supervision of a teacher model that guides the feature distillation process. The joint optimization method rasterizes both RGB images and feature maps simultaneously, ensuring high fidelity and per-pixel accuracy. The optimizable attributes of the $i$-th 3D Gaussian $\Psi_i$ are:

$$
\begin{aligned}
\quad\quad\quad\quad\quad\quad\Psi_i = \left\{ y_i, q_i, s_i, \alpha_i, c_i, f_i \right\},\quad\quad\quad\quad\quad\quad \text{(1)}
\end{aligned}
$$

\noindent where $y_i \in \mathbb{R}^3$ represents the 3D position, $q_i \in \mathbb{R}^4$ denotes the rotation quaternion, $s_i \in \mathbb{R}$ is the scaling factor, $\alpha_i \in \mathbb{R}$ is the opacity value, $c_i \in \mathbb{R}^3$ represents the diffuse color from Spherical Harmonics (SH), and $f_i \in \mathbb{R}^V$ is the feature embedding from the supervised model $F_t$, where $V$ denotes the dimension of the feature vector. Each Gaussian $\Psi_i$ is positioned at $y_i$, and is associated with a feature vector $f_i$ that encodes local spatial and visual content.

The following equations define the computation of pixel color $C$ and pixel feature $F_r$ during rendering:
$$
\begin{aligned}
\quad\quad\quad\quad
C &= \sum_{i \in \mathcal{N}} c_i \alpha_i T_i, \quad  
F_r = \sum_{i \in \mathcal{N}} f_i \alpha_i T_i.\quad\quad\quad\quad \text{(2)}
\end{aligned}
$$

\noindent where $\mathcal{N}$ denotes the set of overlapping 3D Gaussians for a specified pixel, and $T_i$ represents the transmittance, defined as the cumulative opacity of preceding Gaussians overlapping the given pixel.

To train the 3DGS model on a specific scene with grounded feature maps, we define the loss function $\mathcal{L}_{\text{GS}}$ as follows:
$$
\begin{aligned}
\quad\quad\quad\quad\quad\quad\quad\mathcal{L}_{\text{GS}} = \mathcal{L}_{\text{color}} + \mathcal{L}_{\text{features}}
 \quad\quad\quad\quad\quad\quad\quad \text{(3)}
\end{aligned}
$$

The photometric loss $\mathcal{L}_{\text{color}}$ measures the difference between the ground truth image $I$ and the rendered image $\hat{I}$: $\mathcal{L}_{\text{color}} = (1-\lambda) \mathcal{L}_1(I, \hat{I}) + \lambda \mathcal{L}_{\text{SSIM}}(I, \hat{I})$. The feature loss $\mathcal{L}_{\text{features}}$ enforces consistency between the supervised feature map $F_t(I)$ and the rendered feature map $F_r$: $\mathcal{L}_{\text{features}} = \left\| F_t(I) - F_r \right\|_1$.

%% file: src/03-method.tex
\begin{figure*}[!t]
\includegraphics[width=0.99\textwidth]{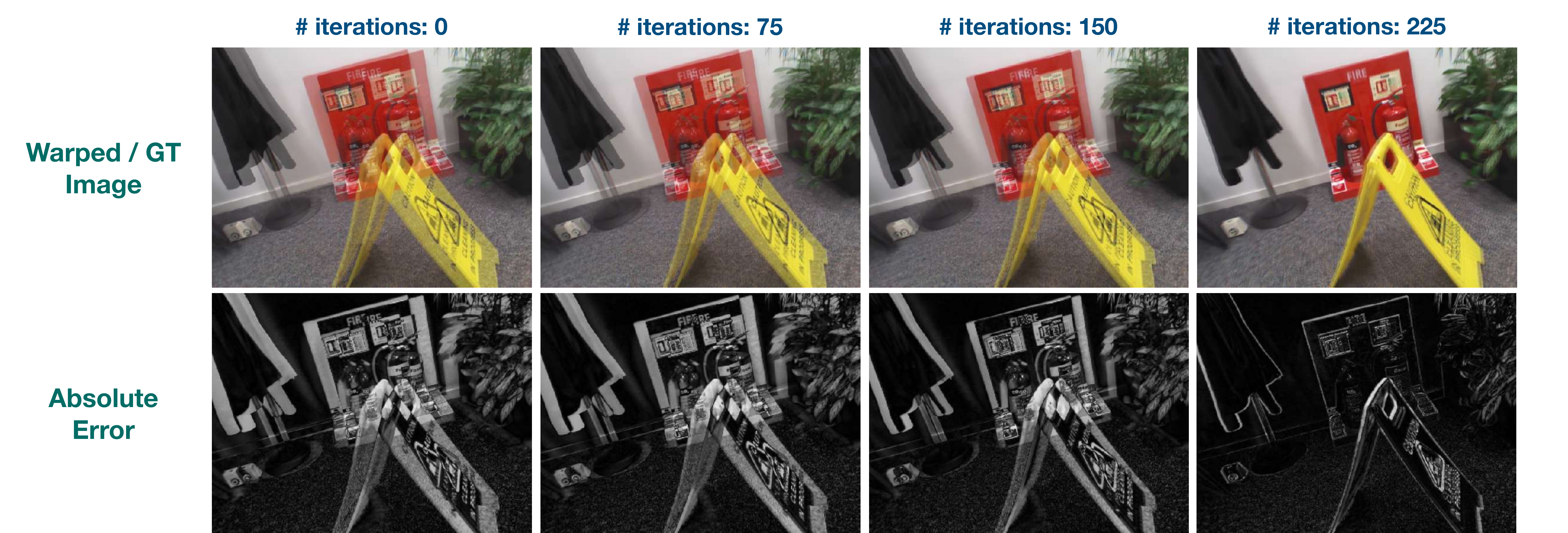}
\caption{\emph{Test-time camera pose refinement} aligns the rendered images to the query image at different optimization iterations. The first row shows the rendered images blended with the query image based on the estimated pose at each step, while the second row visualizes the absolute errors between the two, demonstrating how the warping loss reduces this error over time, thereby improving pose accuracy.}
\label{fig:warp}
\end{figure*}

\section{Methodology}

Our method consists of a two-stage pipeline, as depicted in Figure \ref{fig:pipeline}. The first stage involves modeling the scene, and the second stage focuses on estimating an initial coarse pose, followed by its refinement to improve accuracy.

Initially, we model the scene using a feature-based 3D Gaussian Splatting (3DGS) approach~\cite{zhou2023feature}, guided by a keypoint descriptor network. We use the deep feature extractor XFeat~\cite{potje2024cvpr} due to its robustness in extracting reliable and distinctive features across various environments, for both indoor and outdoor, even in the presence of dynamic elements.

For each training image $ I_t \in \mathbb{R}^{W \times H \times 3} $, the XFeat network computes a feature map~$ F_t(I) \in \mathbb{R}^{(W/8) \times (H/8) \times 64} $, which is bilinearly upsampled to the original resolution. We train the 3D Gaussian Splatting model by minimizing the loss function as described in Equation~3.

Once the scene is learned by 3DGS, we estimate the pose for the query image through a two-phase process: first, estimating a coarse pose, and then refining it.

\noindent\textbf{Obtaining the Initial Coarse Pose}. This stage aims to establish correspondences between 2D keypoints in the query image and the 3D points in the 3DGS model of the scene. We use the Perspective-n-Point (PnP) solver within a RANSAC loop to provide an initial pose estimate.

For a given query image $ q $ accompanied by extracted keypoints $ P_q $ and features $ f_q $ from a descriptor model, we perform 2D-3D correspondence matching with the 3D Gaussian point cloud $ P \in \mathbb{R}^{N \times 3} $ and the associated distilled XFeat features $ f_p \in \mathbb{R}^{N \times 64} $, where $ N $ denotes the number of points. We employ cosine similarity to match the query image 2D features with the 3D scene features distilled in the 3D Gaussians. The 2D-3D correspondences $V(i)$, which represent the matching for the $i$-th pixel, are determined by maximizing the cosine similarity measure:

$$
\begin{aligned}
\quad\quad\quad\quad\quad\quad V(i) = \arg\max_{i \in P} \frac{\mathbf{f}^{i}_{q} \cdot \mathbf{f}^{l}_{p}}{\|\mathbf{f}^{i}_{q}\| \|\mathbf{f}^{l}_{p}\|}\quad\quad\quad\quad\quad\quad \text{(4)}
\end{aligned}
$$

Some methods speed up the search by learning a reliability score for each point. Instead, we use sparse and reliable keypoints from the query image via XFeat and match them with all points in the 3D point cloud, enabling efficient pose estimation through semi-dense matching with the Gaussian cloud of distilled features.

\noindent \textbf{Test-time Camera Pose Refinement.} We enhance the coarse pose estimate in two phases: \emph{optional} feature-based pose refinement and warping-based pose refinement. These methods progressively improve the pose accuracy by minimizing the error between the query and rendered images.
\newline\newline
\noindent\emph{Feature-based Pose Refinement}. The refinement process begins by rendering a feature map based on the coarse pose estimate. In this step, we match keypoints between the query image and the rendered feature map to iteratively refine the pose. Descriptors from the query image are matched with those in the rendered feature map, enabling the identification of corresponding keypoints. These matched keypoints are backprojected into world space using the rendered depth map and the initially estimated pose. This process allows for re-estimation of the pose via the Perspective-n-Point (PnP) algorithm.

This iterative refinement improves pose alignment, with each iteration reducing the error between the query image and the rendered scene. It is particularly effective in complex environments, where even small initial errors can lead to significant misalignments. By refining the pose in this manner, we enhance both the accuracy and robustness of pose estimation, providing more precise results under challenging conditions. However, this step introduces additional computational overhead. Thus, we treat it as optional, depending on the desired trade-off between computation and accuracy. An analysis of this trade-off is provided in the experimental section.
\newline\newline
\noindent\emph{Warping-based Pose Refinement}. In contrast to the feature-based approach, we refine the pose photometrically by aligning the rendered image with the query image through a warping loss. This method is computationally more efficient, requiring only one render pass, compared to iterative rendering in feature-based refinement.

Previous works~\cite{yenchen2021inerf, zhu2022nice} employed gradient descent to minimize photometric residuals, which quantify the difference between the rendered and query images. This process requires neural rendering at each step, making it computationally expensive. PNeRFLoc~\cite{zhao2023pnerfloc} improved this process by introducing a warping loss function. We adopt this warping loss method, which significantly reduces computational cost by rendering the image only once, compared to the repeated rendering required in previous methods. Additionally, the 3DGS framework enhances speed with its faster rendering process compared to NeRF-based methods.

For a query image $q$  and a coarse pose $(\mathbf{R}, \mathbf{t})$, we first render the image $q_r$ and the depth map $d_r$ using the initial pose. We optimize the pose estimate $(\mathbf{R'}, \mathbf{t'})$ by minimizing a warping loss, which is defined as the sum of pixel-wise RGB differences between the reference and query images:

$$
\begin{aligned}
\resizebox{0.9\columnwidth}{!}{$\mathcal{L}_{rgb-warp} = \sum_{p_i} \lVert Y(q, W(p_i, \mathbf{R}, \mathbf{t}, \mathbf{R'}, \mathbf{t'})) - Y(q_r, p_i) \rVert_2$}, \text{(5)}
\label{eq.warp}
\end{aligned}
$$

where
$$
\begin{aligned}
\resizebox{0.9\columnwidth}{!}{$W(p_i, \mathbf{R}, \mathbf{t}, \mathbf{R'}, \mathbf{t'}) = \prod(\mathbf{R}(\mathbf{R'}^{-1}\prod\nolimits^{-1}(p_i, z_r(p_i)) - \mathbf{R'}^{-1}\mathbf{t'}) + \mathbf{t})$}.
\end{aligned}
$$
\newline\newline
Here, $ Y(q_r, p_i) \in \mathbb{R}^{3} $ is the RGB color at pixel $ p_i \in \mathbb{R}^{2} $ on the rendered image $ q_r $. The warp function $ W(\cdot) $ finds the corresponding pixel on the query image $ q $ by warping $ p_i $ from the reference image $ q_r $. More precisely, the function $ W $ back-projects $ p_i $ into the 3D space of $ q_r $’s coordinate system by utilizing the rendered depth $ z_r $, transforms it into the camera coordinate system of $ q $ using the optimized pose $ (\mathbf{R'}, \mathbf{t'}) $, and then projects it onto the image $ q $.

\begin{table*}[!t]
\caption{\emph{Comparison of methods on the 7Scenes dataset}: median translation and rotation errors (cm/°) across various approaches. APR denotes absolute pose regression, SCR represents scene coordinate regression, and NRP stands for neural render pose estimation. The best results are highlighted as \fs{first} and \nd{second}.}
\centering
\setlength{\tabcolsep}{4pt}
\begin{threeparttable}
\resizebox{2\columnwidth}{!}{
\begin{tabular}{l|l|ccccccc|c}
\toprule    & Methods & Chess  & Fire  & Heads & Office &Pumpkin  & Redkitchen& Stairs  & Avg. $\downarrow$ [$\text{cm}/^\circ$] \\
\midrule
 \multirow{4}{*}{APR}& PoseNet~\cite{kendall2015posenet}  & 10/4.02 & 27/10.0& 18/13.0 & 17/5.97& 19/4.67& 22/5.91 & 35/10.5 &  21/7.74\\
 & MS-Transformer~\cite{shavit2021learning}&11/6.38 &23/11.5 &13/13.0 &18/8.14 &17/8.42&  16/8.92& 29/10.3 & 18/9.51 \\
 & DFNet~\cite{chen2022dfnet}&3/1.12 &6/2.30 & 4/2.29& 6/1.54
&7/1.92 & 7/1.74 & 12/2.63 & 6/1.93 \\
 & Marepo~\cite{chen2024maprelativeposeregressionvisual} &1.9/0.83 &2.3/0.92 &2.1/1.24 & 2.9/0.93& 2.5/0.88
& 2.9/0.98 & 5.9/1.48 & 2.9/1.04
 \\\midrule
\multirow{1}{*}{SCR} & ACE~\cite{brachmann2023accelerated}&\nd{0.5}/0.18 &\fs{0.8}/\nd{0.33} & \fs{0.5}/\fs{0.33}& \fs{1.0}/\fs{0.29}& \fs{1}/\fs{0.22} & \fs{0.8}/\fs{0.2} & \fs{2.9}/\fs{0.81} & \fs{1.1}/\fs{0.34} \\
 \midrule
\multirow{6}{*}{NRP}  &FQN-MN~\cite{Germain_2022_CVPR} &4.1/1.31 & 10.5/2.97 & 9.2/2.45 & 3.6/2.36 & 4.6/1.76& 16.1/4.42 & 139.5/34.67 & 28/7.3\\
&CrossFire~\cite{moreau2023crossfirecamerarelocalizationselfsupervised} &1/0.4 &5/1.9 &3/2.3 & 5/1.6 & 3/0.8& 2/0.8 &12/1.9  &4.4/1.38\\
& PNeRFLoc~\cite{zhao2023pnerfloc} &2/0.8 & 2/0.88 &1/0.83 & 3/1.05 & 6/1.51 &5/1.54  & 32/5.73 & 7.28/1.76 \\
& NeRFMatch~\cite{zhou2024nerfectmatchexploringnerf}  &0.9/0.3 & 1.1/0.4 &1.5/1.0 & 3.0/0.8 & 2.2/0.6 &\nd{1.0}/0.3  & 10.1/1.7 & 2.8/0.7 \\
& \textbf{GSplatLoc (Coarse)}  &3.17/0.49 & 3.34/0.7 &1.96/0.76 & 3.8/0.62 & 5.12/0.7 &4.54/0.64  & 10.97/2.63 & 4.7/0.94 \\
& \textbf{GSplatLoc (Base)} &0.43/\nd{0.16} & 1.03/0.32 & 1.06/0.62 & 1.85/0.4 & 1.80/0.35 &2.71/0.55  & 8.83/2.34 & 2.53/0.68 \\
& \textbf{GSplatLoc (Fine)} &\fs{\textbf{0.39}}/\fs{\textbf{0.13}} & \nd{0.91}/\fs{0.29} & \nd{0.94}/\nd{0.50} & \nd{1.41}/\nd{0.32} & \nd{1.41}/\nd{0.26} &1.32/\nd{0.29}  & \nd{3.44}/\nd{0.82} & \nd{1.40}/\nd{0.37} \\
\bottomrule
\end{tabular}
}
\end{threeparttable}
\label{tab:acc_7s}
\end{table*}

%% file: src/04-experiments.tex
\vspace{20pt}
\section{Experiments}

\noindent\textbf{Experimental Setup.} We evaluate our method using the 7Scenes dataset~\cite{glocker2013real-time} for indoor validation and the Cambridge Landmarks dataset~\cite{kendall2015posenet} for outdoor validation.

In the modeling phase, COLMAP~\cite{schoenberger2016sfm} is used to generate point clouds and initialize poses for each scene. We then train the 3D Gaussian Splatting (3DGS) model from~\cite{zhou2023feature} on each scene for 15,000 iterations. XFeat~\cite{potje2024cvpr} is used to extract dense features for 3DGS.

In the testing phase, we start by obtaining an initial coarse pose. We sample 1,000 of the most reliable descriptors and match them to the 3D feature cloud. The number of RANSAC iterations is set to 20,000.

For the refinement step, we render a visual reference for the coarse pose once and use the Adam optimizer with a learning rate of 0.001. We optimize both translation and rotation in quaternion form. 
\newline\newline
\noindent\textbf{GSplatLoc Variants.} We evaluate three variants of the GSplatLoc model:
\begin{itemize}
    \item \emph{GSplatLoc (Coarse)}: Uses only the initial coarse pose estimate.
    \item \emph{GSplatLoc (Base)}: Incorporates photometric warping-based pose refinement for enhanced accuracy.
    \item \emph{GSplatLoc (Fine)}: Combines both feature-based and warping-based refinements for the highest precision.
\end{itemize}
Each variant progressively improves localization accuracy, with trade-offs in computational cost and refinement quality.

Figure~\ref{fig:warp} demonstrates that the warping-based optimization progressively enhances localization accuracy from the initial coarse pose, requiring approximately 250 iterations for indoor scenes and 350 for outdoor scenes, with the visual reference rendered only once. For the feature-based refinement stage, five iterations were selected as the optimal tradeoff between accuracy and efficiency.

\begin{figure}[t]%
\centering
\includegraphics[width=\columnwidth]{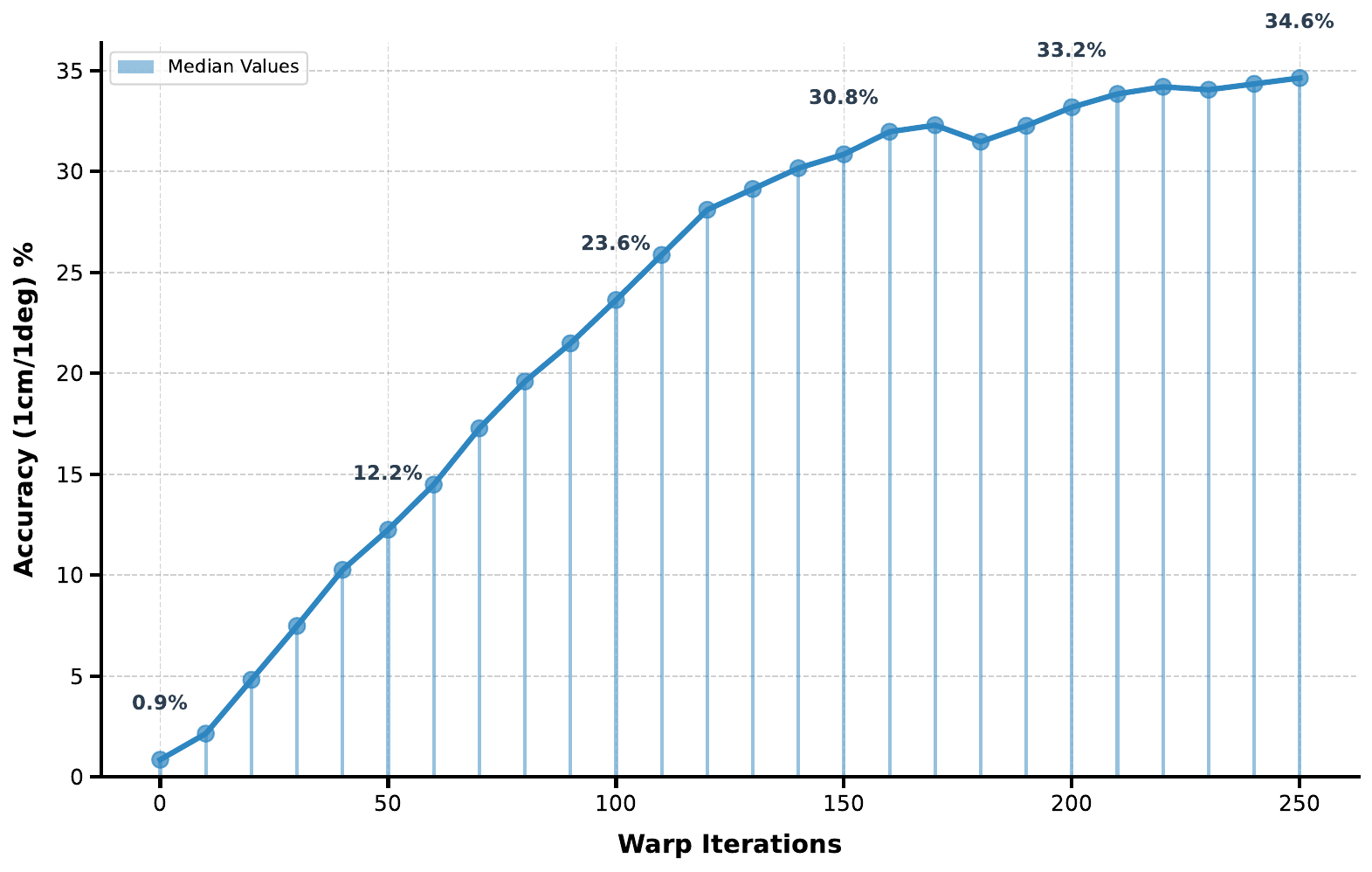}
\caption{\emph{Camera pose optimization via the Base variant} is performed using a rendering-based photometric warp loss, progressively enhancing accuracy. The plot shows the percentage of frames that fall below the $\text{1cm}/1^\circ$ threshold, highlighting the improvement in accuracy over iterations.}
\label{fig:opt}
\end{figure}

Figure \ref{fig:opt} demonstrates the optimization of the camera pose estimated by the Base version of GsplatLoc using a rendering-based photometric warp loss, which iteratively minimizes errors in translation and rotation. Accuracy is evaluated based on the percentage of frames where the pose error is less than 1 cm and 1°. The plot highlights the improvement in accuracy achieved by adding the warp loss.

\begin{table}[!t]
\centering
\caption{\emph{Comparison of methods on the Cambridge Landmarks dataset}: median translation and rotation errors (cm/°) for various approaches. APR denotes absolute pose regression, SCR represents scene coordinate regression, and NRP stands for neural render pose estimation. The best results are highlighted as \fs{first} and \nd{second}.}
\setlength{\tabcolsep}{1pt} %
\begin{threeparttable}
\resizebox{1\columnwidth}{!}{
\begin{tabular}{l|l|cccc|c}
\toprule
 &Methods  & Kings  & Hospital  & Shop & Church  &Avg. $\downarrow$ [$\text{cm}/^\circ$]   \\\midrule
 \multirow{4}{*}{APR}&PoseNet~\cite{kendall2015posenet}  & 93/2.73 & 224/7.88 & 147/6.62 & 237/5.94 & 175/5.79 \\
&MS-Transformer~\cite{shavit2021learning}  &85/1.45 &175/2.43& 88/3.20 & 166/4.12 & 129/2.80 \\
&LENS~\cite{moreau2021lens} &33/0.5 &44/0.9& 27/1.6 & 53/1.6 & 39/1.15 \\
&DFNet~\cite{chen2022dfnet} &73/2.37 & 200/2.98 & 67/2.21 & 137/4.02 & 119/2.90 \\
\midrule
\multirow{1}{*}{SCR}& ACE~\cite{brachmann2023accelerated} &29/0.38 & 31/\nd{0.61} & 5/\nd{0.3} & \nd{19}/0.6 & \nd{21}/\nd{0.47} \\
\midrule
\multirow{5}{*}{NRP}  &FQN-MN~\cite{Germain_2022_CVPR}&28/\nd{0.4} & 54/0.8 & 13/0.6 & 58/2 & 38/1 \\
&CrossFire~\cite{moreau2023crossfirecamerarelocalizationselfsupervised} &47/0.7& 43/0.7 & 20/1.2 & 39/1.4 & 37/1 \\
&PNeRFLoc~\cite{zhao2023pnerfloc} &\fs{24}/\fs{0.29}&\nd{28}/\fs{0.37} & \nd{6}/\fs{0.27} & 40/\nd{0.55} & 24.5/\fs{0.37}\\
& \textbf{GSplatLoc (Coarse)}  &41/0.50&32/0.87 & 11/0.40 & 31/0.72 & 29/0.62\\
&\textbf{GSplatLoc (Base)} &\nd{27}/0.46& 20/0.71 & 5/0.36 & 16/0.61 & 17/0.53\\
&\textbf{GSplatLoc (Fine)} &31/0.49&\fs{16}/0.68 & \fs{4}/0.34 & \fs{14}/\fs{0.42} & \fs{16}/0.48\\
\bottomrule 
\end{tabular}
}
\end{threeparttable}

\label{tab:acc_cam}
\end{table}

\begin{table}[!t]
\centering
\caption{\emph{Comparison of methods on Our Custom Dataset.} The table reports accuracy as the percentage of frames with position and orientation errors below the thresholds ($\text{10cm}/5^\circ$), ($\text{5cm}/5^\circ$), ($\text{2cm}/2^\circ$), and ($\text{1cm}/1^\circ$). It also includes the median translation and rotation errors (cm/°). The best results are highlighted as \fs{first} and \nd{second}.}
\setlength{\tabcolsep}{1pt} %
\begin{threeparttable}
\resizebox{1\columnwidth}{!}{
\begin{tabular}{l|l|cccc|c}
\toprule
 & \shortstack{Method \\ $ $}  & \shortstack{$\text{10cm}/5^\circ$ \\ $\uparrow (\%)$}   & \shortstack{$\text{5cm}/5^\circ$ \\ $\uparrow (\%)$}   & \shortstack{$\text{2cm}/2^\circ$ \\ $\uparrow (\%)$} & \shortstack{$\text{1cm}/1^\circ$ \\ $\uparrow (\%)$}  & \shortstack{Median Error \\ $\downarrow$ (cm/°) }    \\\midrule

\multirow{1}{*}{SCR}& ACE~\cite{brachmann2023accelerated} & \nd{98.6} & \nd{93.9} & \nd{76.9} & 53.1 & 1.0/0.20 \\
\midrule
\multirow{3}{*}{NRP}
& \textbf{GSplatLoc (Coarse)}  & 95.2 &  83.0 & 37.4 & 10.2 & 2.5/0.40\\
&\textbf{GSplatLoc (Base)} & 97.3 & 90.5 & 73.5 & \nd{57.8} & \nd{0.8}/\nd{0.12} \\
&\textbf{GSplatLoc (Fine)} & \fs{99.3} & \fs{99.3} & \fs{90.5} & \fs{78.2} & \fs{0.4}/\fs{0.06} \\
\bottomrule
\end{tabular}
}
\end{threeparttable}
\label{tab:acc_custom}
\end{table}

\begin{table*}[!t]
\caption{\emph{Effect of feature descriptors.} The table reports the accuracy metric as the percentage of frames below ($\text{1cm}/1^\circ$)/( $\text{5cm}/5^\circ$) position and orientation discrepancies thresholds, descriptor dimension, and size per scene in MB. Tested on the 7Scenes dataset. The best results are highlighted in \fs{first} and \nd{second}.}
\centering
\setlength{\tabcolsep}{4pt}
\begin{threeparttable}
\resizebox{2\columnwidth}{!}{
\begin{tabular}{l|c|c|ccccccc|c}
\toprule  Backbone  & Dim.$\downarrow$& Size(MB)$\downarrow$ & Chess(\%)  & Fire (\%)& Heads(\%) & Office(\%) &Pumpkin(\%)  & Redkitchen(\%)& Stairs (\%)  & Avg.(\%) $\uparrow$ \\
\midrule
DeDoDe-G  & \nd{256} & \nd{$\sim$800} & 37.1/94.2 & 17.2/69.8 & \nd{33.1/86.8} & 1.8/50.9 & 1.6/38.5  & 4.7/48.7 & 0.6/20.5 & 13.7/58.5 \\
SuperPoint  & \nd{256} & \nd{$\sim$800} & \nd{70.9/99.7} & \nd{38.2}/\fs{93.0} & 25.9/76.9 & \fs{19.7/88.0} & \nd{13.2}/\fs{81.2}  & \fs{27.3/82.8} & \fs{5.1/56.2} & \nd{28.6}/\fs{82.56} \\
XFeat  & \fs{64} & \fs{$\sim$300} & \fs{81.2/99.8} & \fs{46.8}/\nd{88.3} & \fs{51.6/87.6} & \nd{16.2/83.1} & \fs{18.5}/\nd{78.2}  & \nd{26.9/77.4} & \nd{1.2/31.8} & \fs{34.6}/\nd{78.0} \\
\bottomrule
\end{tabular}
}
\end{threeparttable}
\label{tab:ablation_backbones}
\end{table*}

\noindent\textbf{Performance Results.} We compare our results with several state-of-the-art methods and report the median translation and rotation errors in (cm/degree) for each scene and the average across all scenes in Tables~\ref{tab:acc_7s} and \ref{tab:acc_cam}.


For the indoor 7Scenes dataset, our method, GSplatLoc, surpasses the previous state-of-the-art neural render pose estimation method, NeRFMatch, in six out of seven scenes, setting a new state-of-the-art among NRP methods. However, GSplatLoc is still outperformed by ACE, a scene coordinate regression approach.

\begin{figure}[!t]
\includegraphics[width=0.5\textwidth]{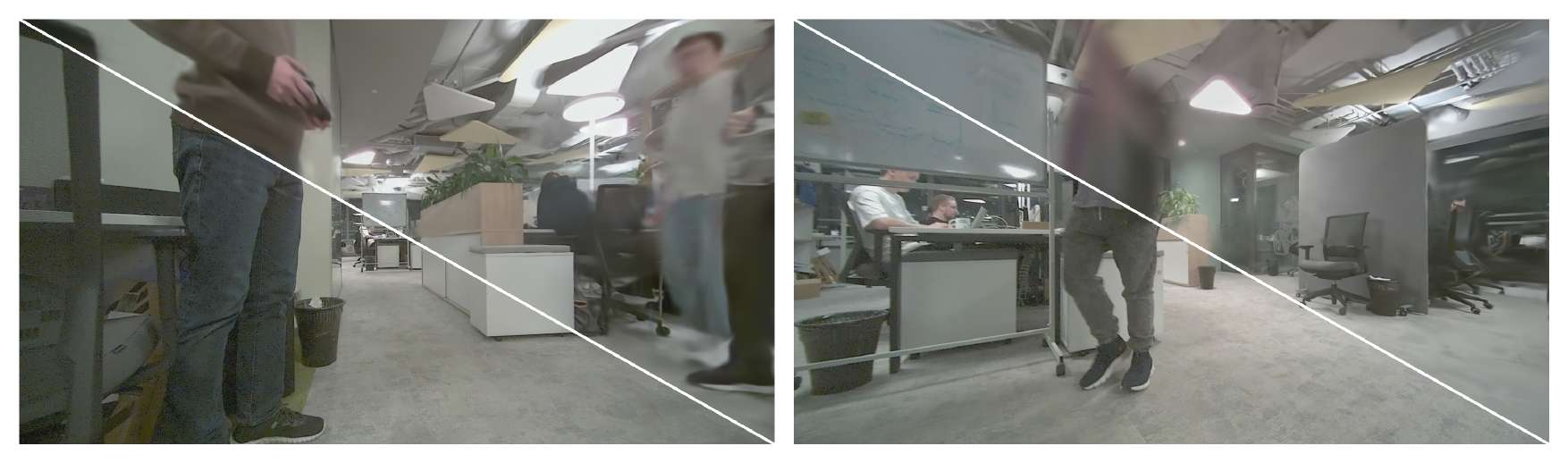}\caption{\emph{Qualitative results on Our Custom Dataset.} The diagonal line separates the test query images from the renders synthesized using poses estimated by \emph{GSplatLoc Fine}.}
\label{fig:real_world}
\end{figure}

On the outdoor dynamic Cambridge Landmarks dataset, GSplatLoc achieves the best overall performance, surpassing ACE—the second-best method—by an average of more than 5 cm in translation error. This emphasizes that the proposed method excels in handling dynamic scenarios by incorporating a robust outlier rejection step,

Furthermore, we position our solution as a downstream approach for robot localization. To validate this capability, we built a custom dataset captured in a dynamic indoor environment. Specifically, a quadruped robot equipped with a Livox Mid-360 LiDAR and a ZED X camera navigated through an office-like area filled with moving people. The environment also includes numerous transparent objects such as mirrors, which present significant challenges for many localization techniques. Some qualitative results of this custom dataset can be seen in figure \ref{fig:real_world}. Instead of relying on COLMAP, we leveraged LiDAR priors for scene reconstruction. Each scene was optimized for 15,000 iterations. During inference, we used 250 iterations for warping-based refinement and 5 iterations for feature-based pose refinement. In this challenging scenario, our method outperforms the state-of-the-art indoor relocalization approach (ACE). The corresponding results are presented in Table~\ref{tab:acc_custom}.

\begin{table}[!t]
\centering
\caption{\emph{Runtime analysis} of the execution time per a query frame for each method across different stages.}
\label{tab:ablation_time}
\setlength{\tabcolsep}{3pt} 
\begin{threeparttable}
\resizebox{1\columnwidth}{!}{
\begin{tabular}{l | c c c c}
\toprule
\shortstack{Method\\ $ $}&\shortstack{Feature Est.\\ (s)}&\shortstack{Rendering\\ (s)}&\shortstack{Refinement\\ (s)}&\shortstack{Overall Query\\ (s)}   \\
\midrule
PNeRFLoc    & N/A      & N/A      & N/A & 5.560 \\
NeRFMatch    & 0.157  & 0.141 & 0.846 & 1.144  \\
\textbf{GSplatLoc (Base)} \,  & 0.018  & 0.140  & 0.651 & 0.809 \\
\textbf{GSplatLoc (Fine)} \,  & 0.018  & 0.140  & 1.911 & 2.069 \\
\bottomrule
\end{tabular}
}
\end{threeparttable}
\end{table}

\noindent\textbf{Effect of Feature Descriptors.} We evaluated XFeat~\cite{potje2024cvpr}, DeDoDe-G~\cite{Edstedt2023DeDoDeDD}, and SuperPoint~\cite{detone2018superpoint} in the GSplatLoc Base setting to assess the impact of different feature descriptors. Results in Table~\ref{tab:ablation_backbones} show that XFeat, the lightest, outperforms DeDoDe-G and matches SuperPoint at the $\text{5cm}/5^\circ$ threshold, while surpassing both at the $\text{1cm}/1^\circ$ threshold. XFeat provides a higher percentage of frames within the tighter accuracy threshold but has more "outlier frames", slightly reducing its accuracy compared to SuperPoint. Additionally, XFeat requires less storage, making it a more efficient choice.

\noindent\textbf{Runtime Analysis.} Table~\ref{tab:ablation_time} shows the execution time of our method on the 7Scenes dataset. The shortest processing time is required for coarse pose search and rendering, while the longest time is taken by the iterative process of refining with warp loss over 250 iterations. NeRFMatch's refinement time is computed by averaging across six iterations, with each optimization step taking 141 ms~\cite{zhou2024nerfectmatchexploringnerf}.

Our pipeline outperforms NRP methods by balancing execution speed and estimation accuracy. NeRFMatch and PNeRFLoc require several seconds per query, whereas our method delivers a coarse pose estimate in just 0.2 seconds. 
On an RTX 4090 GPU, our method achieves superior accuracy compared to competitors in under 1 second.

\section{Discussion}
Our approach outperforms other neural rendering pose estimation (NRP) methods, achieving the best overall accuracy in indoor settings. For outdoor settings, which involve dynamic objects and lighting variations, we demonstrate superior position estimation accuracy. Although \mbox{PNeRFLoc}~\cite{zhao2023pnerfloc} achieves higher accuracy on certain sequences, its performance significantly degrades on others. This variability shows that our method offers more robust performance and greater reliability when transitioning between indoor and outdoor environments, making it well-suited for diverse use cases, such as delivery robots and AR applications.

Iterative optimization is crucial for accurate localization, but poor reconstruction can degrade warp loss and photometric optimization. In the Stairs scene from the 7Scenes dataset, the Fine version of GSplatLoc, which includes feature-metric pose optimization, outperforms the Base version. This underscores the importance of feature-based optimization for reliable localization, though it comes with a tradeoff between accuracy and computational efficiency.

Additionally, the use of 3DGS for spatial representations significantly accelerates the training process compared to NeRF-based NRP methods, playing a key role in scaling for large outdoor scenes.
Another strong competitor, ACE~\cite{brachmann2023accelerated}, being better indoors, it struggles with outdoor scenarios, since its shallow MLP-based scene encoding design constrains its modeling capabilities.

At the same time, NRP methods, to which GSplatLoc belongs, are a better alternative, since it is a more versatile framework than SCR, as it allows solving multiple tasks in parallel using the same scene representation. For example, 3DGS can be used to encode semantic or language-aligned instances or even model dynamic scenes—tasks that are crucial for robotics. This flexibility enables a broader range of interactions with the environment and enhances localization. 

We explain the overall good performance of GSplatLoc by two main reasons. First, feature distillation combined with 3DGS enables accurate structure-based coarse pose estimation. Second, representations that provide realistic images facilitate effective photometric optimization.

%% file: src/05-conclusion.tex
\section{Conclusions and Future Work}

We present GSplatLoc, a framework using 3D Gaussian Splatting (3DGS) for visual localization in indoor and outdoor environments, including dynamic scenes. The method employs a two-stage process: a coarse pose estimate from 2D-3D correspondences, followed by warping loss refinement. By combining structure-based matching with rendering optimization from scene-agnostic 3DGS descriptors, we improve accuracy and efficiency. 
Our approach outperforms APR and NRP methods indoors and surpasses SCR-based ACE outdoors, handling dynamic objects and lighting while being the fastest among NRP methods. We also demonstrate the effectiveness of the lightweight XFeat feature extractor. Future work will focus on removing floaters from the 3DGS model, extending it to large-scale outdoor scenarios, and applying NRP for semantic SLAM and navigation.